\documentclass[runningheads]{llncs}
\usepackage{graphicx}
\usepackage[colorlinks]{hyperref}
\hypersetup{colorlinks, linkcolor=green,citecolor=green}
\usepackage{algorithm}
\usepackage{algorithmic}
\usepackage{graphbox}
\usepackage{enumerate}
\usepackage{amsmath,amssymb}

\newcommand\blfootnote[1]{%
  \begingroup
  \renewcommand\thefootnote{}\footnote{#1}%
  \addtocounter{footnote}{-1}%
  \endgroup
}

\begin{document}
\title{Surgical Instruction Generation with Transformers}

\author{Jinglu Zhang\inst{1}\and
Yinyu Nie\inst{2,*} \and
Jian Chang\inst{1}\and
Jian Jun Zhang\inst{1}
}

\authorrunning{J. Zhang et al.}

%
\institute{National Centre for Computer Animation (NCCA), Bournemouth University, UK \and Technical University of Munich}

\maketitle              

\begin{abstract}
Automatic surgical instruction generation is a prerequisite towards intra-operative context-aware surgical assistance. However, generating instructions from surgical scenes is challenging, as it requires jointly understanding the surgical activity of current view and modelling relationships between visual information and textual description. Inspired by the neural machine translation and imaging captioning tasks in open domain, we introduce a transformer-backboned encoder-decoder network with self-critical reinforcement learning to generate instructions from surgical images. We evaluate the effectiveness of our method on DAISI dataset, which includes 290 procedures from various medical disciplines. Our approach outperforms the existing baseline over all caption evaluation metrics. The results demonstrate the benefits of the encoder-decoder structure backboned by transformer in handling multimodal context.
\end{abstract}
\blfootnote{*Corresponding email: yinyu.nie@tum.de}
\section{Introduction}
Surgical instruction generation is a task of automatically generating a natural language sentence to guide surgeons of how to perform the operation based on the current surgical view. It is an essential component towards building context-aware surgical system, which aims to utilize available information inside the operation room to provide clinicians with contextual support at appropriate time. Moreover, when on-site mentoring is unavailable or a rare case is detected, providing intra-operative surgical instructions by expert surgeons is imperative. However, surgical data has high heterogeneity even for the same type of surgery due to different surgical skill level, medical condition, and patient specific situation. Accordingly, understanding surgical content and generating a natural language sentence to guide the procedure is challenging.

Previously, telementoring~\cite{bilgic2017effectiveness}, which exchanges medical information through video and audio in real time, has been proved as an efficient solution for intra-operative guidance, including pointing out target anatomical structure from the monitor, controlling the camera or the robotic arm, etc. Nonetheless, telementoring is limited by the cost of specific equipment and software, the high demand of transport speed, and legal and ethic issues~\cite{bilgic2017effectiveness,erridge2019telementoring}. With the huge development of related techniques of context-aware surgical assistance, understanding and analyzing the surgical activities inside the operation room opens up the possibility of providing intra-operative assistance for surgeons. Most of the existing researches focus on surgical phases and fine-grained gestures recognition~\cite{twinanda2016endonet,funke2019using,zhang2020symmetric}. However, these methods can be regarded as the segmentation and classification problems based on pre-defined phases and gestures, thus have no ability of generating the unseen instructions.  

The most related research topic to us is medical report generation~\cite{jing2017automatic,chen2020generating,bustos2020padchest}, which describes the \textbf{\textit{impression}}, \textbf{\textit{findings}}, \textbf{\textit{tags}}, etc. of a patient in reference to the radiology or pathology. One of the earliest medical report generation works based on natural language is~\cite{jing2017automatic}, which jointly predicts tags and generates long paragraphs with co-attention and hierarchical LSTM. More recently,~\cite{chen2020generating} improves the transformer model~\cite{vaswani2017attention} by designing a relational memory to record key information of the generation process and provides a memory-driven layer normalization for transformer decoder. Despite the challenges, medical reports also have their own discriminating characters. They often share predefined topics and follow similar writing templates, while surgical instruction generation with natural language has no template to follow. 

To our best knowledge,~\cite{rojas2020daisi} is the only prior work for surgical instruction generation. In their work, the authors create the Database for AI Surgical Instruction dataset (DAISI) and use a bidirectional recurrent neural network (RNN) to generate the description for a surgical image. However their work has two limitations. For one, although RNNs are designed for sequence generation with arbitrary length, they suffer from the essential vanishing gradient problem~\cite{pascanu2013difficulty}. For another, they apply the BLEU~\cite{papineni2002bleu} score as the only evaluation metric, which is insufficient for natural language evaluation. 

In this paper, inspired by the great performance of transformer model in machine translation~\cite{vaswani2017attention} and image captioning~\cite{cornia2020meshed} from the open domain, we build our network with an encoder-decoder fully backboned transformer to generate surgical instructions. Taking an surgical image as the input, we first extract its visual attention features by a fine-tuned ResNet-101 module. Then the encoder attention blocks, decoder attention blocks, and encoder-decoder attention blocks model the dependencies for visual features, textual features, and visual-textural relational features, respectively. On the other hand, sequence generation models are often trained using the cross-entropy (XE) loss and evaluated using non-differential metrics such as BLEU, CIDEr~\cite{vedantam2015cider}, etc. In order to alleviate the mismatch between training and testing and improve the evaluation performance, we apply the reinforcement learning based self-critical approach~\cite{rennie2017self} to directly optimize the CIDEr score. Experimentally, we extensively explore the performance of different baselines (LSTM-based fully connected and soft-attention models) on DAISI dataset~\cite{rojas2020daisi}. The experiments demonstrate that our transformer-backboned architecture outperforms the existing methods as well as our other proposed baselines. The promising instructions generated from the network bring potential value in clinical practice.
\section{Methodology}
In this section, we introduce our framework in details. It involves two sub-modules: 1) the transformer-backboned encoder-decoder structure for surgical instruction generation (see section~\ref{EDTrans}); 2) the self-critical reinforcement learning for optimizing the CIDEr score (see section~\ref{RL}).

\subsection{Encoder-Decoder with Transformer Backbone}\label{EDTrans}
\begin{figure}[htb!]
\centering
\includegraphics[width=0.7\textwidth]{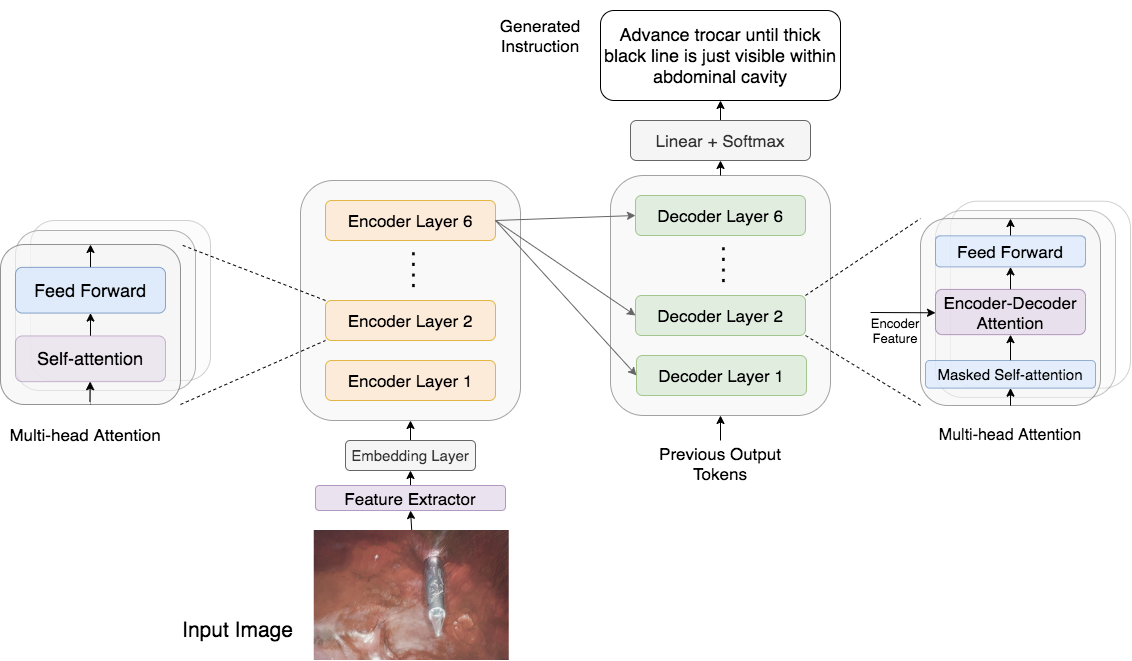}
\caption{The transformer-backboned surgical instruction generation architecture} 
\label{figure_transformer}
\end{figure}

The whole encoder-decoder structure can be seen in Figure~\ref{figure_transformer}. Following the modern learning paradigm, we design this kind of structure to encode latent features from images and decode them into natural languages. Before our network, a ResNet-101~\cite{he2016deep} is adopted to output $14\times14\times2048$ image features, which are afterwards embeded by a linear embedding layer to reduce the dimension to $14\times14\times512$ followed by a ReLU and a dropout layer. Subsequently, our encoder firstly processes the flattened spatial features ($196\times512$) and produces non-local relationships between image regions. Then the decoder takes hidden attentive representation from the encoder outputs and generates the corresponding instruction with natural language. The essential attention mechanism behind the transformer is called scaled dot-product attention~\cite{vaswani2017attention}, which is defined as:
\begin{equation}
    \mathrm{Attention} (Q,K,V) = \mathrm{Softmax} (\frac{QK^T}{\sqrt{d}})V
\end{equation}
where $Q$ is the packed query matrix, $K$ and $V$ are packed key-value pairs, and $d$ is a scaling factor (equals to the dimension of $K$). It calculates a weighted sum of the values based on the similarity distribution between the query with all the keys. 

The whole encoder is a stack of $6$ attention blocks with identical structure. Specifically, each block consists of a \textbf{multi-head self-attention} layer (8 heads) and a position-wise \textbf{feed-forward network}. The multi-head self-attention layer is represented as:
\begin{equation}
\begin{aligned}
    MultiHead(Q,K,V)&= Concat(head_1,...,head_h)W^O\\
    \mathrm{head_i}& = \mathrm{Attention}(Q{W^Q_i}, K{W^K_i}, V{W^V_i})
\end{aligned}
\end{equation}
where matrices ${W^Q_i}$, ${W^Q_i}$, ${W^Q_i}$ and $W^O$ are projection parameters to be learned during the training phase. Different linear transformations are applied to the queries, values, and keys for each attention head. A simple position-wise fully connected feed-forward network is then applied to each attention layer:
\begin{equation}
    \mathrm{FFN(x)} = \mathrm{max}(0,xW_1 + b_1)W_2 + b_2,
\end{equation}
where $W_1$,$W_2$ and $b_1$,$b_2$ are corresponding weights and biases for two fully connected layers. 

The input of the decoder is the hidden representation exported from the last encoder layer. The decoder also consists of six identical blocks, where each has two multi-head attention layers (a decoder self-attention layer and an encoder-decoder attention layer) and one fully connected feed-forward network. Every decoder self-attention layer is masked to prevent from attending to future locations. For further explanation of the decoder, please refer to the original transformer paper.\cite{vaswani2017attention}.
\subsection{Reinforcement Learning}\label{RL}
Sequence generation models are often trained in "Teacher-Forcing"~\cite{bengio2015scheduled} mode, which inputs the ground-truth to maximize the likelihood of next prediction during training and uses previously generated words from the model distribution to predict the next word during test time. In order to bridge this gap, we apply the self-critical reinforcement learning as proposed in~\cite{rennie2017self}. After pre-training the model with standard word-level XE loss, the CIDEr score~\cite{vedantam2015cider} is directly optimized as the reward. All the detailed formula derivation can be found in~\cite{rennie2017self}. 
\section{Evaluation}
\subsection{Experimental Settings}
 \noindent\textbf{Dataset Description.} We evaluate our approach on DAISI dataset~\cite{rojas2020daisi}, which contains 17,255 color images from 290 medical procedures, including external fetal monitoring, laparoscopic sleeve gastrectomy, laparoscopic ventral hernia repair, etc. The availability of the dataset is upon request\footnote{https://engineering.purdue.edu/starproj/}. Every procedure consists of few images with their corresponding instruction texts. We further clean the dataset by deleting noisy and irrelevant images and descriptions. Finally, there are 16,413 images in total with one caption for each image. While some surgical procedures have only one sample due to the limited dataset size, we split the data in per image manner. We assign 13,094 images for training, 1,646 for validation, and 1,673 for testing.

 \noindent\textbf{Text Preprocessing.} Text preprocessing is a significant step to transform the text into a more analyzable and predictable format for the deep learning model. Raw text instructions need to be preprocessed to learn meaningful features and not overfit on irrelevant noise. We follow these steps to clean the text instruction: 1) Converting all words to lower case; 2) Expanding abbreviations, including medical abbreviations (e.g. `a.' to `artery') and English contractions (e.g. i've to `i have'); 3) Removing numbers, punctuation, and whitespace; 4) Tokenizing the sentence into words. 
 
We further set the threshold of the sentence length to 16, label 
any word count less than five as 'UNK', and build a vocabulary of size 2212 words. 

\noindent\textbf{Evaluation Metrics.} Besides the instruction generation task, how to automatically evaluate the generated sentences has become increasingly important. The key idea is to measure the correlation of generated captions with human judgments. Following most of the image captioning methods, we apply 1-4 gram BLEU~\cite{papineni2002bleu}, Rouge-L~\cite{lin2004rouge}, METEOR~\cite{banerjee2005meteor}, CIDEr~\cite{vedantam2015cider}, and SPICE~\cite{anderson2016spice} to evaluate our model, while the first three metrics are originated from machine translation and the last two metrics are specifically designed for image captioning.

\subsection{Implementation and Training Details}
All the models are implemented in PyTorch and trained on a single NVIDIA GeForce GTX 1080 graphics card. We first train our model with a word-level cross-entropy (XE) loss, then optimize the model using reinforcement learning. It takes around 30 hours for the training process (30 epochs for general XE loss, and 30 epochs for reinforcement learning). During the XE training process, the model is trained to predict the next word given previous ground-truth word, while the reinforcement learning process is trained to predict next word based on the previous prediction. It takes around 30 hours for the training process (30 epochs for general XE loss, and 30 epochs for reinforcement learning)

\noindent\textbf{Transformer Encoder-Decoder.} We use ResNet-101~\cite{he2016deep} pre-trained on ImageNet classification task to extract image features. A spatially adaptive max-pooling layer is applied after the final convolution layer. It ends up with a fixed size of $14\times 14 \times 2048$-d (196 image regions in total) output. For the XE training, we initialize the learning rate to $3 \times 10^{-4}$ and follow the learning rate scheduling strategy with 20000 warm-up steps for 30 epochs. During the self-critical evaluation, we use a fixed learning rate of $1 \times 10^{-5}$ for another 30 epochs. Both models are optimized using ADAM optimizer~\cite{kingma2014adam} with a batch size of 5.

\noindent\textbf{LSTM-based Models.}
In order to comprehensively evaluate the surgical instruction generation task, we implement two additional models for comparison and discussion, namely LSTM model and LSTM-based soft-attention model similarly to~\cite{xu2015show,vinyals2015show}. For LSTM model, images are encoded to 2048 dimension vectors with the final convolution layer of ResNet-101 followed by an average pooling layer. The LSTM-based soft-attention model shares the same image feature maps with transformer model. For both models, the image embedding, words embedding dimension and LSTM hidden state size are set to 512. 
\section{Results and Discussion}
\subsection{Comparison with the State-of-the-Art}
We clean the original dataset~\cite{rojas2020daisi} by removing noisy and wrong image-text pairs. Thus a new benchmark is required. As the code in~\cite{rojas2020daisi} is not publicly available, we re-implement their Bi-RNN model. The 4096 dimensional image features are extracted using the last convolutional layer from a pre-trained VGG16~\cite{simonyan2014very}. The Bi-RNN model is trained with 50 epochs by the initial learning rate at $5 \times 10^{-4}$ and the batch size at 10. Table~\ref{table_daisi} compares our proposed models with \cite{rojas2020daisi}, which shows that Bi-RNN has relatively lower performance, especially for the 3-gram and 4-gram BLEU scores (11.3\% and 9.3\%) compared with ours (46.4\% and 44.9\%). In BLEU score evaluation, long $n-gram$ score measures the fluency of the instruction. It can be concluded that Bi-RNN is not capable of generating "human-like" instructions. 

\begin{figure}[htb!]
\centering
\includegraphics[width=0.75\textwidth]{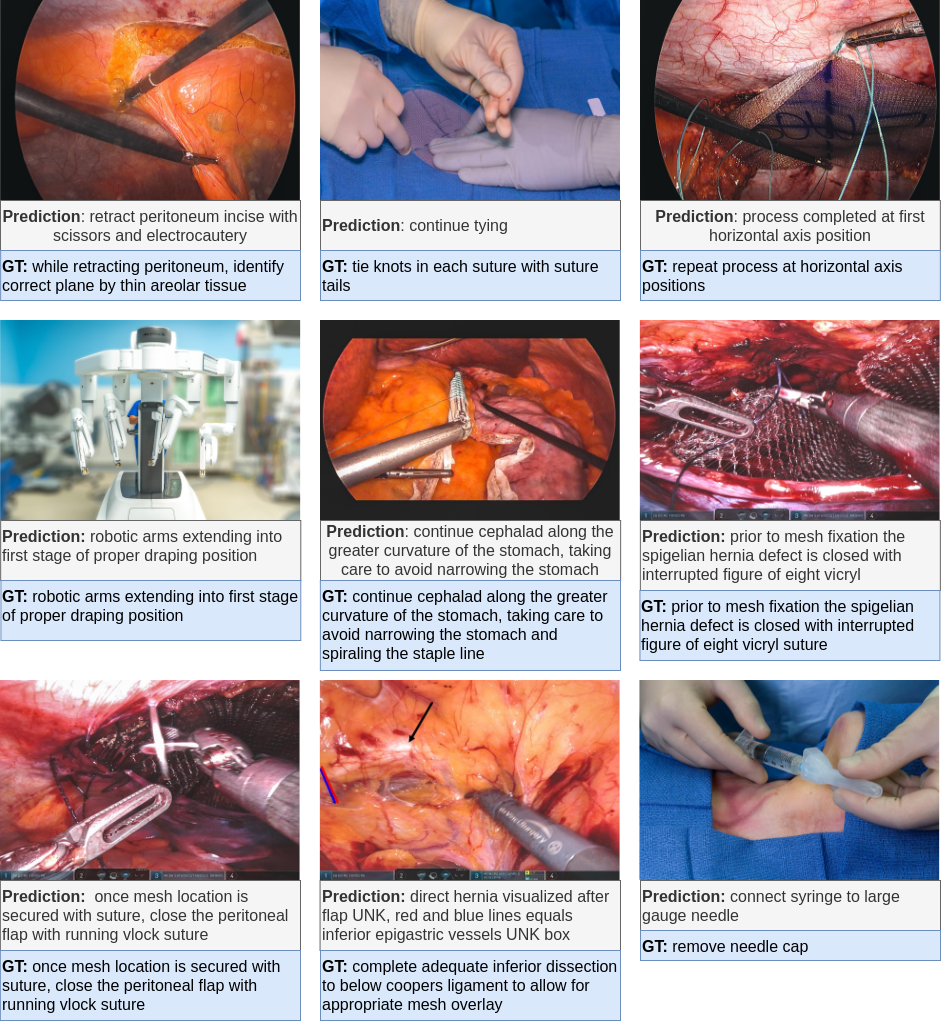}
\caption{Qualitative results with transformer-backboned encoder-decoder framework.} 
\label{figure_vis}
\end{figure}

LSTM model achieves slightly better performance than LSTM-based soft-attention approach, and the transformer model outperforms all the others in all metrics. This indicates that the conventional RNN-based methods have limited ability of catching the dependencies between image features and text information. While transformer-backboned encoder-decoder layers can encode the dependencies for image pixels, the  self-attention layers in decoder are able to model dependencies for textual information, and the encoder-decoder attention builds the relationship between image and textual features. Figure~\ref{figure_vis} shows some visualization samples using the proposed transformer-backboned framework.

\begin{table*}[htb!]
\centering
\caption{Comparison with the state-of-the-art~\cite{rojas2020daisi} for surgical instruction generation task. B1, B2, B3, B4, C, M, R and S stands for 1-4 gram BLEU, CIDEr, METEOR, ROUGE-L and SPICE score respectively.}
\begin{tabular}{lllllllll}
\hline
\begin{tabular}[c]{@{}c@{}}Surgical\\ Instruction\end{tabular} & \textbf{$B1$} & \textbf{$B2$} & \textbf{$B3$} & \textbf{$B4$} & \textbf{$C$} &\textbf{$M$} &\textbf{$R$} &\textbf{$S$}\\
\hline
DAISI (Bi-RNN) & 21.0 & 14.4 & 11.3 & 9.3 & 8.32 & 10.3 & 22.0 & 12.1\\
LSTM  & 43.7 & 39.4 & 37.3 & 36.2 & 34.0 & 24.9 & 44.6 & 40.2 \\
LSTM + soft-attn & 43.2 & 38.7 & 36.3 & 34.9 & 32.4 & 24.3 & 43.7 & 38.0 \\
Transformer + rl & \textbf{52.8} & \textbf{48.7} & \textbf{46.4} & \textbf{44.9} & \textbf{42.7} & \textbf{30.7} & \textbf{53.1} & \textbf{48.4} \\
\hline
\end{tabular}
\label{table_daisi}
\end{table*}

\subsection{Effects of Reinforcement Learning}
To further explore the functionality of each design in our network, we decouple three networks and design an ablative experiment in six settings: (1) LSTM only; (2) LSTM + reinforcement learning; (3) LSTM + soft-attention; (4) LSTM + soft-attention + reinforcement learning; (5) Transformer only; (6) Transformer + reinforcement learning. The experiment results are shown in Table~\ref{table_rl}.

\noindent\textbf{(1) v.s. (3):} We add the soft-attention module on the top of LSTM to sequentially attend to different parts of image and aggregate information, but it performs slightly worse (around 1\% for each evaluation standard) than the baseline model. This indicates that simple soft-attention mechanism cannot build the correlation between salient pixels and the next word prediction.

\noindent\textbf{(1) v.s. (3) v.s. (5):} Without using any recurrent neural units as LSTM-based models, transformer-backboned model only use the attention mechanism to encode the image information and decode its corresponding text instruction. Transformer-backboned model achieves better performance than two LSTM models, which demonstrate its ability in handling multi-modal contexts.

\noindent\textbf{(1) v.s. (2), (3) v.s. (4), and (5) v.s. (6):} During the training procedure, we first train each model with standard XE loss, then we add the reinforcement learning block to optimize the CIDEr score directly. From the results, it can be seen that not only the CIDEr score, but also the performance of other evaluation metrics has been lifted. Specifically, we observe a significant increasing when using reinforcement training after the transformer-backboned model. 

\begin{table*}[htb!]
\centering
\caption{Ablative study to explore the influence of reinforcement learning. B1, B2, B3, B4, C, M, R and S stands for 1-4 gram BLEU, CIDEr, METEOR, ROUGE-L and SPICE score respectively.}
\begin{tabular}{lllllllll}
\hline
\begin{tabular}[c]{@{}c@{}}Surgical\\ Instruction\end{tabular} & \textbf{$B1$} & \textbf{$B2$} & \textbf{$B3$} & \textbf{$B4$} & \textbf{$C$} &\textbf{$M$} &\textbf{$R$} &\textbf{$S$}\\
\hline
LSTM  & 43.7 & 39.4 & 37.3 & 36.2 & 34.0 & 24.9 & 44.6 & 40.2 \\
LSTM + rl & 44.6 & 40.3 & 38.3 & 37.1 & 35.1 & 25.4 & 45.3 & 41.1 \\
LSTM + attn & 43.2 & 38.7 & 36.3 & 34.9 & 32.4 & 24.3 & 43.7 & 38.0 \\
LSTM + attn + rl & 43.4 & 38.8 & 36.4 & 34.8 & 33.1 & 24.8 & 44.1 & 38.5 \\
Transformer & 45.5 & 41 & 38.7 & 37.2 & 34 & 25.6 & 44.3 & 39.7 \\
Transformer + rl & \textbf{52.8} & \textbf{48.7} & \textbf{46.4} & \textbf{44.9} & \textbf{42.7} & \textbf{30.7} & \textbf{53.1} & \textbf{48.4} \\
\hline
\end{tabular}
\label{table_rl}
\end{table*}

\subsection{Limitations and Challenges}
In this part, we discuss the current challenges and limitations for automatic surgical instruction generation. 
\begin{enumerate}
    \item \textbf{Small dataset size.} Deep learning algorithms often require huge amount of data to tune the parameters and prevent overfitting, e.g., COCO dataset~\cite{lin2014microsoft} has more than 120K samples. Surgical instruction generation is a multi-modal problem, which relates visual, text, and the relationship between them. Therefore, the solution space is much larger than other tasks (e.g. classification and segmentation). However excluding the noisy and irrelevant images, DAISI dataset only contains 16,413 images. 
    \item \textbf{No fine-grained supervisions.} In feature extraction, some image captioning algorithms use Faster R-CNN algorithm~\cite{ren2016faster} to detect object bounding boxes and identify attribute features with Visual-Genome data~\cite{anderson2018bottom}. However, obtaining semantic and attributive annotations in medical science is quite challenging since it requires expert annotators.
    \item \textbf{One caption per image.} In real life, an image can be described in different ways. For example, COCO captioning task has equipped with 5 different reference translations for each image. Nonetheless, we have only one annotation per image. It is possible that the evaluation metrics grade an adequate caption a low score only because it does not fit the ground truth.
\end{enumerate}
\section{Conclusion}
In this paper, we propose an encoder-decoder architecture fully backboned by transformer to generate surgical instructions from various medical disciplines. The experiment results demonstrate that transformer architecture is capable of creating the pixel-wise patterns from self-attention encoder, developing text relationships for masked self-attention decoder, and devising the image-text dependencies from encoder-decoder attention. In order to solve the mismatching between the training and testing procedure, we optimize the model with self-critical reinforcement learning, which takes the CIDEr score as the reward after the general cross-entropy training. 

Understanding surgical activity and generating instruction is still at its early stage. Future works include collecting the large training dataset, building the specialized pre-trained model for medical images, regularizing and annotating more reference captions for surgical images.
\bibliographystyle{splncs04}
\bibliography{reference.bib}
\end{document}